\documentclass[preprint,12pt,authoryear]{elsarticle}

\usepackage{amssymb}

\usepackage{graphicx}
\usepackage{bm}
\usepackage{makecell}
\usepackage{multirow}
\usepackage{collcell}
\usepackage{colortbl}
\usepackage{booktabs}
\usepackage{url}
\usepackage{graphicx}
\usepackage{xcolor}

\newcommand{\TrainingSet}{\bm{T}} 
\newcommand{\TestSet}{\Psi}

\begin{document}

\begin{frontmatter}



\title{Enhancing Coronary Artery Calcium Scoring via Multi-Organ Segmentation on Non-Contrast Cardiac Computed Tomography}


\author[2,1]{Jakub Nalepa}
\ead{jakub.nalepa@polsl.pl}
\author[1]{Tomasz Bartczak}
\ead{tbartczak@graylight-imaging.com}
\author[1]{Mariusz Bujny\corref{cor1}}
\ead{mbujny@graylight-imaging.com}
\author[1]{Jarosław Gośliński}
\ead{jgoslinski@graylight-imaging.com}
\author[1,3]{Katarzyna Jesionek}
\ead{katarzyna.jesionek@us.edu.pl}
\author[1]{Wojciech Malara}
\ead{wmalara@graylight-imaging.com}
\author[1]{Filip Malawski}
\ead{fmalawski@graylight-imaging.com}
\author[1,4]{Karol Miszalski-Jamka}
\ead{k.miszalski-jamka@sccs.pl}
\author[1]{Patrycja Rewa}
\ead{prewa@graylight-imaging.com}
\author[1,3]{Marcin Kostur}
\ead{marcin.kostur@us.edu.pl}

\cortext[cor1]{Corresponding author, e-mail: mbujny@graylight-imaging.com}

\affiliation[1]{organization={Graylight Imaging},
            addressline={ul. Bojkowska 37a}, 
            city={Gliwice},
            postcode={44-100}, 
            country={Poland}}

\affiliation[2]{organization={Silesian University of Technology},
            addressline={ul. Akademicka 16}, 
            city={Gliwice},
            postcode={44-100}, 
            country={Poland}}

\affiliation[3]{organization={University of Silesia},
            addressline={ul. Bankowa 12}, 
            city={Katowice},
            postcode={40-007}, 
            country={Poland}}

\affiliation[4]{organization={Silesian Center for Heart Diseases},
            addressline={ul. Marii Skłodowskiej-Curie 9}, 
            city={Zabrze},
            postcode={41-800}, 
            country={Poland}}

\begin{abstract}
Despite coronary artery calcium scoring being considered a largely solved problem within the realm of medical artificial intelligence, this paper argues that significant improvements can still be made. By shifting the focus from pathology detection to a deeper understanding of anatomy, the novel algorithm proposed in the paper both achieves high accuracy in coronary artery calcium scoring and offers enhanced interpretability of the results. This approach not only aids in the precise quantification of calcifications in coronary arteries, but also provides valuable insights into the underlying anatomical structures. Through this anatomically-informed methodology, the paper shows how a nuanced understanding of the heart's anatomy can lead to more accurate and interpretable results in the field of cardiovascular health. We demonstrate the superior accuracy of the proposed method by evaluating it on an open-source multi-vendor dataset, where we obtain results at the inter-observer level, surpassing the current state of the art. Finally, the qualitative analyses show the practical value of the algorithm in such tasks as labeling coronary artery calcifications, identifying aortic calcifications, and filtering out false positive detections due to noise.
\end{abstract}




\end{frontmatter}


\section{Introduction}

Coronary artery disease (CAD) is one of the leading causes of death worldwide~\citep{Kwiendacz2023}. The adopted predictive measurement for assessing the CAD risk is coronary artery calcium (CAC) scoring~\citep{doi:10.1148/ryct.2021200484}. Here, non-contrast (NC) computed tomography (CT) scans are used to evaluate the patient condition in terms of calcifications in coronary arteries (CA). However, manual analysis of 3D CT images is time-consuming and prone to human bias, thus automatic analysis is crucial for providing reproducible and objective diagnosis on a large scale. There are three main CAC scoring methods, including the Agatston, volume, and mass scores~\citep{BLAHA2017923}. All of them rely on voxel-level recognition of calcifications in the CT scans, which is a semantic segmentation problem. The current medical standard states that only objects with value $\geq130 \mathrm{HU}$ should be considered as calcifications~\citep{https://doi.org/10.1002/acm2.12806}. This threshold very well distinguishes calcifications and soft tissues. However, hard tissues (such as bones) and image acquisition noise may have high HU values too~\citep{TEAGUE2012232}. Thus, it is pivotal to distinguish between such true and false positive detections which would affect the extracted scores. Although the total CAC score for the CA is a relevant risk indicator, a more precise diagnosis can be obtained by identifying calcifications in four different regions of the CA tree. Two main vessels stem from the aorta---the left coronary artery (LCA) and the right coronary artery (RCA). RCA is considered as a single region, while LCA includes three ones: the left main coronary artery (LM), left circumflex artery (LCx), and the left anterior descending artery (LAD). Both RCA and LM areas are adjacent to the aorta, and therefore an important challenge is to distinguish which calcifications, or parts of calcifications, belong to the aorta and which to those regions. Hence, we also address this problem in the approach proposed in this paper.


\subsection{Related Work}
Since the introduction of the orCaScore challenge at the MICCAI 2014 \citep{https://doi.org/10.1118/1.4945696}, there has been a notable surge in interest toward employing machine learning (ML) for the task of CAC scoring. However, two primary issues persist: (\textit{i})~distinguishing coronary from non-coronary calcifications and image artifacts, and (\textit{ii})~analyzing each coronary calcifications on a per-vessel basis, by attributing each calcification to a specific coronary artery. In calcification identification, a range of heuristics~\citep{7493209}, classic methods~\citep{https://doi.org/10.1118/1.4945045}, and deep learning were used~\citep{10.1117/12.2293681,gogin,zhang}. Moreover, to eliminate the confusion with non-coronary calcifications, a pre-defined region of interest around the heart is employed to prune false positives~\citep{https://doi.org/10.1118/1.4945045}. Calcification localization is often approached through the use of black-box models that produce multi-class calcification segmentation maps~\citep{zhang,hong,yu}, where the assignment of the calcification to a vessel cannot be explained to the clinician. Alternatively, accurate segmentation of coronary arteries is deemed essential, yet, delineating coronary arteries in NC CTs is challenging. Some techniques offer to produce a total Agatston score, limiting their explainability~\citep{gogin,ihdayhid}. There are methodologies leveraging both contrast-enhanced and NC CTs, where the former assist in organ segmentation~\citep{https://doi.org/10.1118/1.4945045}. Another perspective emphasizes the estimation of CAC using non-ECG-gated CTs~\citep{yu,suh}. A more recent work by \cite{takahashi2023fully}, which relies on the segmentation of anatomical regions rather than on end-to-end identification of calcifications based on the image, stresses the need to build and employ ML models for accurate localization of coronary arteries in NC CT during the calcium scoring process. According to the authors, such an approach would make ML-based calcium scoring explainable, and therefore, would be an ideal method, but it is challenging due to poor visibility of coronary arteries in NC CT. Precise localization of coronary arteries would most probably also help to eliminate false positive detections present in their approach, associated with calcifications of, e.g., pericardium, pulmonary trunk, mediastinal lymph node, and mitral annulus.

While many solutions have been proposed for automating CAC scoring in NC CTs~\citep{zhang}, the common missing factor in state-of-the-art methods is the interpretability of their results. Segmentation and labeling of calcifications is typically performed using different heuristics, mostly based on the distance to heart chambers. While on average such methods produce good results, it is very difficult to identify when they fail, as the medical expert is provided only with numerical results and calcifications overlaid on CT scans, without any anatomical rationale. Since coronary arteries are poorly visible in NC CTs, it is extremely demanding for clinicians to verify if the segmented calcifications are correct, particularly in low-quality examinations. We tackle these shortcomings in this work.

\subsection{Contribution}

We propose an interpretable method for the segmentation and labeling of calcifications, which takes advantage of the novel framework for segmentation of CA in NC CT scans, despite their poor visibility. With our approach, we are able to provide the clinicians with reasonable estimation of CA vessel tree, which allows for instantaneous visual evaluation and interpretation of the segmented calcifications. We argue that even though CA segmentations are often fragmented, they provide sufficient information for medical experts to improve the diagnosis process, particularly for difficult cases. Thus, our contributions are multi-fold:

\begin{itemize}
    \item We propose an end-to-end anatomically-informed approach for the automatic segmentation and classification of calcifications (according to their anatomical locations), and for extracting the CAC score (Section~\ref{sec:method}).
    \item We show that our pipeline extracts interpretable maps of calcifications labeled with their anatomical location which may enhance the analysis.
    \item We rigorously verify our methodology over more than 500 heterogeneous non-contrast ECG-gated cardiac CT scans acquired using multiple CT scanner models, and show that the obtained results are at the inter-observer accuracy level (Section~\ref{sec:experiments}).
    \item We demonstrate the superior accuracy of the method by competing in the orCaScore Grand Challenge, where our approach is currently the best (Section~\ref{sec:experiments}).
\end{itemize}

\noindent We believe that our efforts may be a step toward building reproducible, objective, and interpretable artificial intelligence pipelines enhancing the management of patients with cardiovascular diseases, ultimately leading to personalized care.


\section{Materials and Methods}

In Section~\ref{sec:dataset}, the dataset used in this study is presented. Section~\ref{sec:method} discusses our approach for extracting more interpretable calcium score information thanks to the calcifications' segmentation and their localization within the heart's anatomy. 

\subsection{Dataset and Ground-Truth Generation}
\label{sec:dataset}

We use a dataset of 534 NC CTs, of which 71 came from the orCaScore challenge~\citep{https://doi.org/10.1118/1.4945696}, and the remaining CTs were acquired in three clinical centers from two European countries. It was split into training ($\TrainingSet$) and test ($\TestSet$) subsets, with $\TrainingSet$ of 412 scans (16 from orCaScore) and $\TestSet$ of 122 scans (55 from orCaScore). To capture similar patient distributions in $\TrainingSet$ and $\TestSet$, we stratified these sets according to five risk categories of total calcium scores ($<$1, 1--10, 11--100, 101--400, $>$400). Of note, $\TestSet$ covered a range of scanner manufacturers: GE (18 scans), Philips (17), Siemens (68), and Cannon (19), to capture real-world data heterogeneity.

As we develop an approach for precise calcium scoring in CA with the assignment of scores to four CA territories (RCA, LM, LAD, LCx), the ground truth (GT) with labeled calcifications was meticulously prepared. First, calcification proposals were obtained using automated thresholding (130 HU) and the distance to CA. Then, 4 qualified medical experts (2--4 years of experience, YOE) reviewed the results and manually annotated calcifications which were missed in the previous step using the K3D-jupyter (v. 2.15.2) and Blender (v. 2.91.0) tools. Next, the calcifications were labeled as belonging to one of the four regions of the CA tree. The main difficulty here was concerned with the division of calcifications that were single components extending to multiple areas. To prepare precise GT for CA, calcifications located in coronary ostia were split into aortic and CA parts, and only the latter were included in the final GT. Each GT delineation was reviewed by an experienced cardiologist (17 YOE).

\subsection{Interpretable Calcification Segmentation Enhances Calcium Scoring}\label{sec:method}

\begin{figure}[ht!]
    \centering
    \includegraphics[scale=1.0]{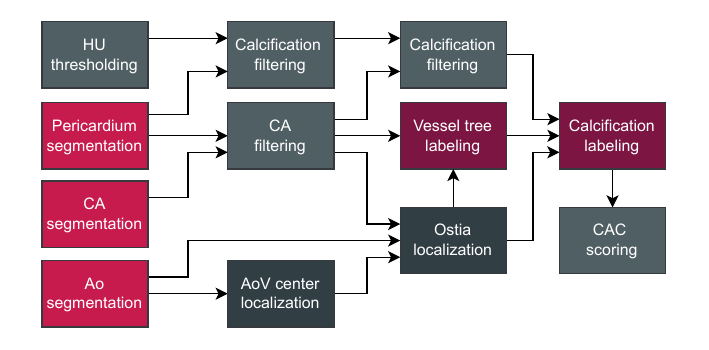}
    \caption{Overview of the proposed method. The colors correspond to specific types of modules (segmentation, localization, labeling, filtering and scoring).}
    \label{fig:pipeline}
\end{figure}

\begin{figure}[ht!]
    \centering
    \includegraphics[width=\textwidth]{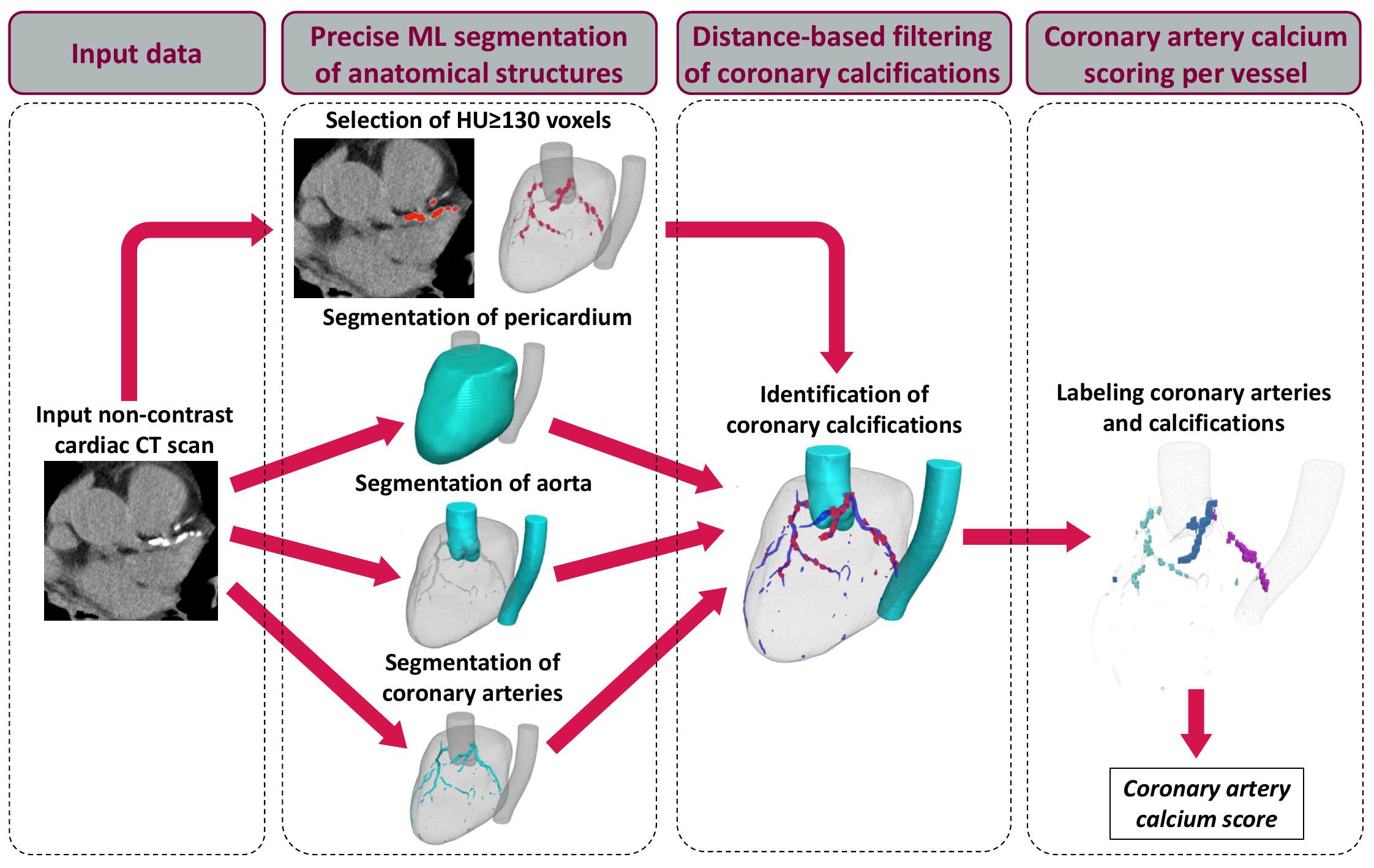}
    \caption{Graphical overview of the main steps of the proposed algorithm.}
    \label{fig:graphical_overview}
\end{figure}

\begin{figure}[ht!]
\centering
\newcommand{\mywidth}{0.475}
\begin{tabular}{cc}
    (a) & (b) \\
    \includegraphics[width=\mywidth\textwidth]{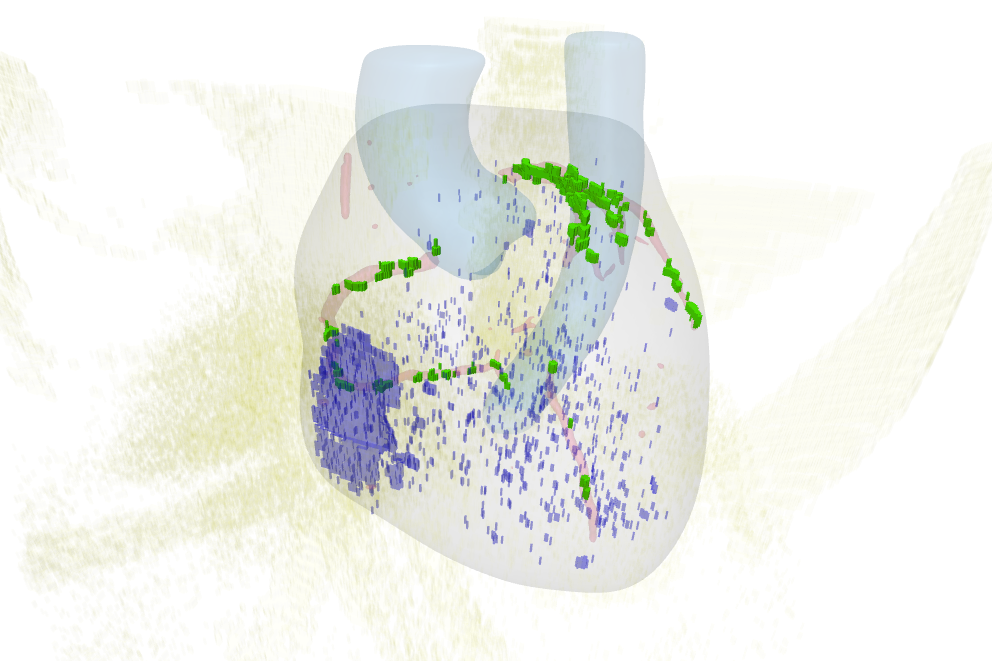}& \includegraphics[width=\mywidth\textwidth]{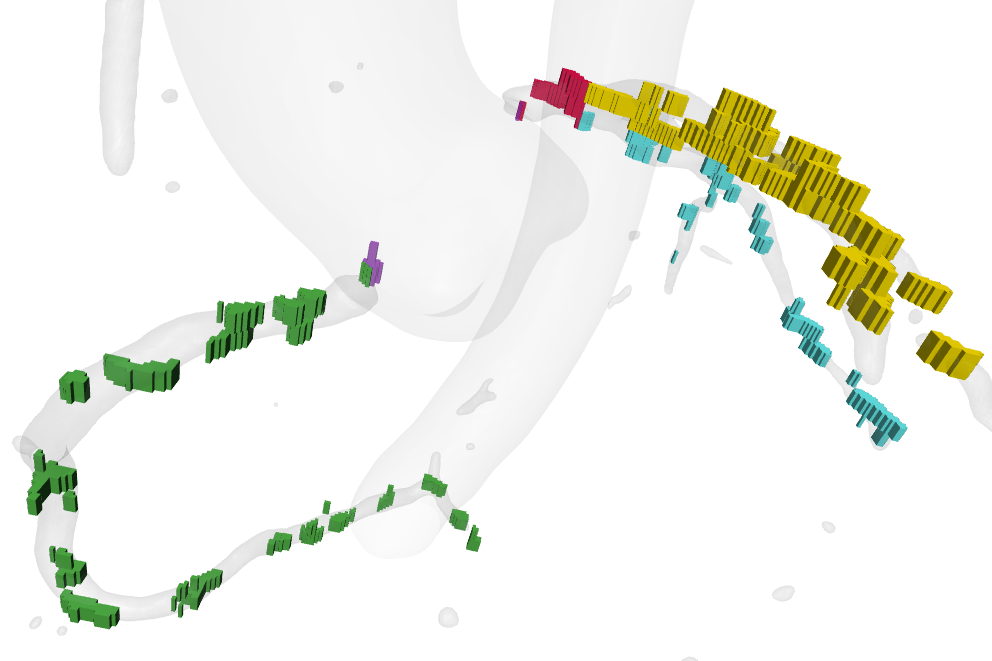}
\end{tabular}
\caption{Visual summary of our method to obtain labeled CA calcifications: (a)~three levels of calcification proposals: thresholding at 130 HU (yellow, translucent voxels), filtering using pericardium segmentation mask (blue, translucent voxels), filtering using CA segmentation mask (green voxels), and (b)~labeling of calcifications based on their proximity to the CA areas (green---RCA, red---LM, yellow---LAD, cyan---LCx) and removal of aortic calcifications (purple, translucent voxels).}\label{fig:visualalgo}
\end{figure}

Our approach includes several pivotal steps: (\textit{i})~segmentation of the scans, (\textit{ii})~localization of calcifications and coronary ostia, (\textit{iii})~their filtering and (\textit{iv})~labeling, and ultimately (\textit{v}) extracting quantifiable calcification scores. A detailed summary of the proposed method is provided in Fig.~\ref{fig:pipeline}, while Fig.~\ref{fig:graphical_overview} shows a graphical overview of the main steps of the algorithm. The input 3D CT scan is processed to extract calcification candidates via thresholding, as defined by medical standards. Here, we also elaborate segmentation masks of pericardium, CA and aorta, using deep learning models. Next, calcification candidates and CA are filtered by removing objects positioned outside the dilated pericardium segmentation mask, which removes significant part of false positive detections. Meanwhile, the aorta segmentation is used for localization of the aortic valve center, which is used, along with the aorta segmentation mask and the filtered CA segmentation mask, as input to ostia localization. Vessel tree labeling is performed using both filtered CA mask and ostia points. Calcification candidates are filtered using the information on the distance to the filtered CA mask, and then labeled based on the vessel tree labeling and coronary ostia localization. Finally, the quantifiable measures are extracted from the resulting calcifications. The next sections present these steps in more detail, and Fig.~\ref{fig:visualalgo} renders a visual summary of our method.

\subsubsection{Segmentation of Cardiovascular Structures}
Obtaining segmentation masks for relevant cardiovascular structures allows for precise and interpretable detection of calcifications. For the pericardium, aorta \citep{bujny_seeing_2024} and CA segmentation, we exploit the training procedure and nnU-Net models, as suggested by \cite{bujny2024coronary}, using a semi-manually labeled CTs. Since CA are poorly visible in NC CTs, creating manual GT would be very challenging, even for experienced experts, as it would require the corresponding reference from contrast-enhanced images. Thus, we employed registration of high-resolution coronary tree models obtained in contrast CT with NC CTs.

\subsubsection{Localization of Calcifications}
Locating calcifications within the regions of the CA tree is of high clinical relevance. An important step of this process is identifying the CA inflows (medically called \textit{ostia}), being the locations in which the aorta turns into arteries. Since the CA segmentations in NC are usually fragmented, relying on the closest points between CA and aorta masks is not sufficient. Instead, we start with a rough localization of the aortic valve center, by using a convolutional neural network (CNN) with an additional dense layer as a regressor, which operates solely on downsampled ($2\times$) aorta mask. Then, we refine this localization with another CNN, operating on a limited area ($67\times67\times67$\,mm) around the initially estimated point, but using the original aorta segmentation rather than the downsampled one. According to our preliminary experiments, the two-step process is both more precise and faster (due to downsampling) than using a single stage network. The third CNN determines the locations of both LCA and RCA ostia in an area around the estimation of the aortic valve center, using both aorta and CA segmentation masks. The architecture of the employed CNNs corresponds to the encoder of the U-Net~\citep{10.1007/978-3-319-24574-4_28}, only with an additional dense layer of 6 neurons (one for each of the three coordinates of the two ostia).

\subsubsection{Filtering of False Positive Calcification Candidates}
By medical definition, calcifications are considered only in voxels with HU$\geq 130$. Thus, calcification candidates are obtained using simple thresholding. However, it results in a significant number of false positives. We use a dilated pericardium mask (with the experimentally fine-tuned dilation size of 1\,mm) to remove all detections positioned outside the heart. Similarly, this step is applied to CA. Finally, calcifications are filtered by investigating their distance to the coronary vessels. This is consistent with the medical perspective, as clinicians evaluate calcifications in each coronary vessel. The CA mask is first dilated by 3\,mm to compensate for segmentation inaccuracy, and then all candidates not intersecting with the dilated CA are removed.

\subsubsection{Labeling of Calcifications}

The calcification labeling includes: (\textit{i})~identifying which calcifications, or parts of calcifications, belong to the aorta, and (\textit{ii})~assigning the remaining calcifications to specific branches of the vessel tree. It is common for the calcifications to be located in the coronary ostia region, either as lesions in the CA and aorta, or as a single lesion extending to both regions. To split such calcifications, we need to find the boundary between aorta and CA. While this is not a well-defined anatomical structure, we estimate it based on the medical suggestions. We process the left and right coronary artery separately, using the determined ostia center points. For each point, we consider the surrounding patch of size $25\times25\times25$\,mm, and compute the Euclidean distance transform (EDT) for both aorta and CA. Then, we create a separation plane by maximizing distance to both masks.

Calcifications (or their parts) on either side of the separation plane are assigned to respective regions---the aorta or CA. CA calcifications are assigned to one of four regions: RCA, LM, LAD, LCx, based on the distance to the vessel tree regions, which are determined as follows. The inflows of the LCA and RCA branches are identified using the located ostia center points. Then, LCA and RCA masks are processed by connecting disjoined components corresponding to the RCA and LCA branches. RCA is considered as a separate region, hence it requires no further processing. LCA is skeletonized~\citep{lee94}, and converted to a directed acyclic graph, in which a node is a $n$-furcation of the vessel tree, and the edge represents a part of the vessel between consecutive $n$-furcations. The vessel morphology, such as its diameter, is represented as graph's edge attributes. The logistic regression classifier is trained on manually engineered features, including the relative angle between edges, thickness and segment's length, to recognize subregions of LCA at each node of the graph. Starting from the inflow, the graph is traversed until all subregions are identified. The RCA and LCA trees are finally used to label calcifications by the closest distance.

\section{Experimental Study}\label{sec:experiments}

\begin{figure}[ht!]
\newcommand{\mywidth}{0.495}
\scriptsize
\centering
\setlength{\tabcolsep}{3.1pt}
\scalebox{0.9}{\begin{tabular}{cc}
\multicolumn{2}{c}{(a)}\\ \\
    \multicolumn{2}{c}{\begin{tabular}{rcccccccccccc}
\Xhline{2\arrayrulewidth}
& &  \multicolumn{3}{c}{DICE} & & \multicolumn{3}{c}{Sensitivity}&&\multicolumn{3}{c}{Specificity}\\
\cline{2-5} \cline{7-9}\cline{11-13}
 Model (GT)   &   & Mean& Weighted  & Count  &   & Mean & Weighted  & Count  &   & Mean & Weighted  & Count \\
 \hline
Ours (our GT)&&0.983&\textbf{0.987}&0.986&&
0.981&0.982&\textbf{0.981}&&0.988&\textbf{0.993}&0.992\\
Ours (our orCaScore GT)&&0.984&0.982&0.981&&
0.989&\textbf{0.987}&\textbf{0.986}&&0.980&0.978&0.977\\
Ours (orCaScore GC GT)&&---&0.980&---&&
---&0.974&0.927&&---&0.986&0.971\\

\hline
{Observer 1}&& --- &0.986&---&&
---&0.985&0.943&&---&0.988&0.993\\
\hline
{3$^{\rm rd}$ place}&& --- &0.979&---&&
---&0.970&0.767&&---&0.988&0.979\\

\Xhline{2\arrayrulewidth}
\end{tabular}} \end{tabular}}  \\[1cm]

\scalebox{0.9}{
\begin{tabular}{c}
  (b) \\ 
  \includegraphics[width=0.75\textwidth]{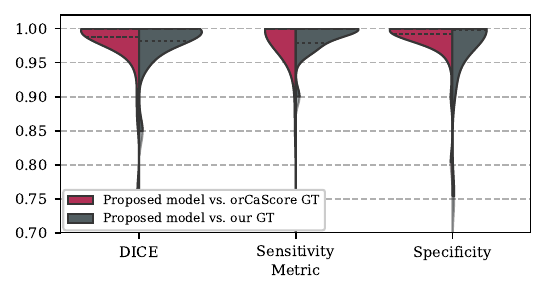} \\
  (c) \\ 
  \includegraphics[width=0.75\textwidth]{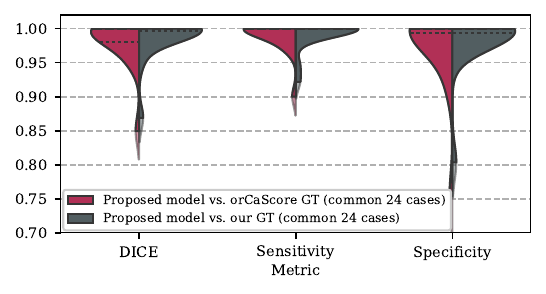}
\end{tabular}}

\caption{The experimental results (a) DICE, sensitivity and specificity---mean, and the mean weighted by their volume in mm$^3$ (Weighted), and by the number of calcifications in GT (Count), with (b)~distribution of metrics computed on complete sets of GT, and (c)~distribution of metrics for CTs where both types of GT (our and orCaScore) are available. The metrics that outperform {Observer 1} are \textbf{bold}. }\label{table:results}
\end{figure}

The objectives of our experiments are two-fold: (\textit{i})~to verify the calcification segmentation performance, and (\textit{ii})~to investigate the agreement between the Agatston score (AS) extracted manually and automatically for all CA areas. The segmentation quality is quantified using DICE, sensitivity (Sen) and specificity (Spe)---all metrics should be maximized ($\uparrow$) and range from 0 to 1. The Intraclass Correlation Coefficient (ICC) was calculated on a single measurement, absolute-agreement, two-way random-effects model to verify the agreement between expert and automatic AS. We conducted two types of evaluations---the first was performed using our test set (Sect.~\ref{sec:dataset}), including expert GT annotations for four labels: LM, LAD, LCx, and RCA. Here, the mean and the mean weighted by both the total number and volume of lesions were calculated. To confront our approach with the state of the art, we conducted another analysis using data from the orCaScore Grand Challenge (GC). This analysis was carried out in two different ways: (\textit{i})~for a subset of the orCaScore cases from our test set, we calculated all relevant metrics and their distributions based on the GT calcification segmentations by our medical imaging experts; (\textit{ii})~for the 40 test cases used in the orCaScore GC for evaluating the algorithms, we performed an automated analysis with the method proposed in the paper, and made an official submission in the system to compute the metrics. In both cases, to compare with the published results, we applied a minor modification of the model's predictions---the calcifications labeled as LM were changed to LAD. The experiments were run on an NVIDIA A100 40 GB GPU, and our pipeline offers fast operation, with the average end-to-end inference time of 6 min.

\begin{figure}[ht!]
    \centering
    \newcommand{\mywidth}{0.5}
    \scalebox{0.96}{\begin{tabular}{cc}
        (a) & (b) \\
         \includegraphics[width=\mywidth\textwidth]{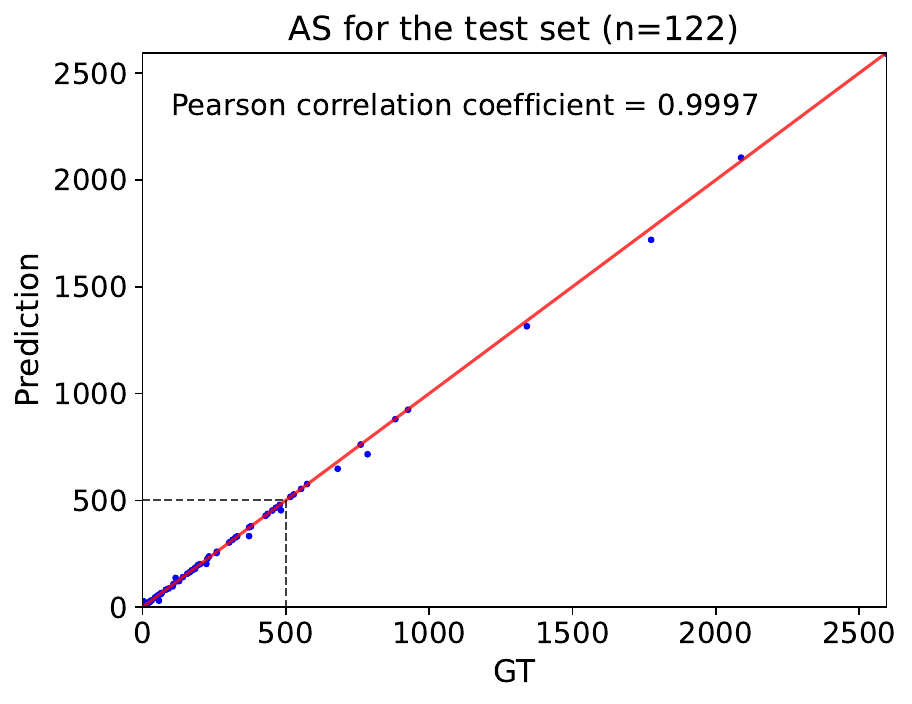}   & \includegraphics[width=\mywidth\textwidth]{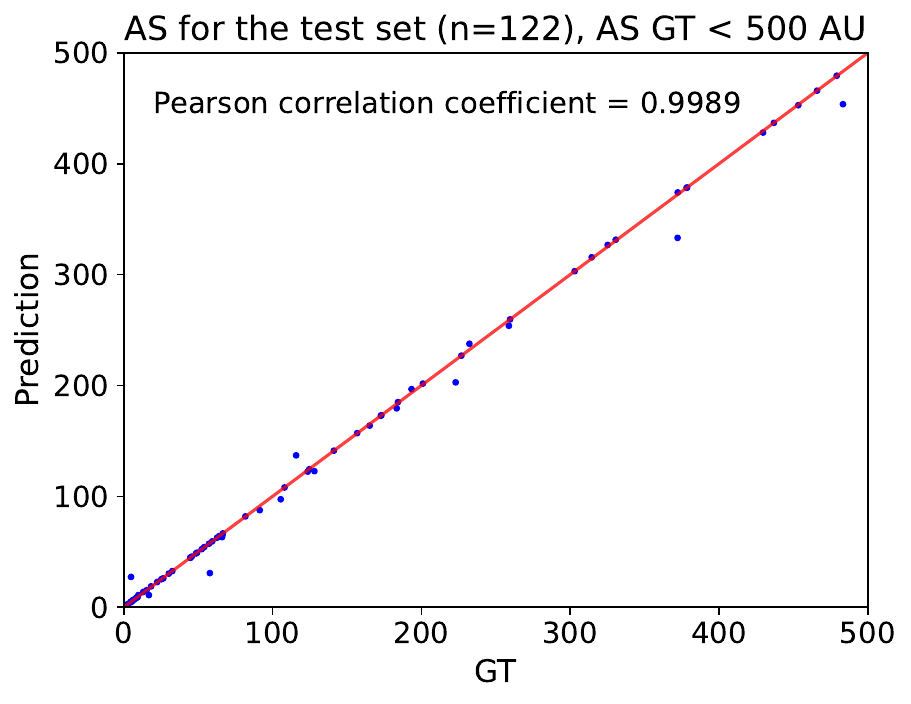}
    \end{tabular}}
    \caption{Agatston score (AS) correlation plot for our test set: (a) for all samples, (b) for AS $<$ 500 Agatston units (AU).}
    \label{fig:as_corr_plot}
\end{figure}

\begin{table}[ht!]
\scriptsize
\centering
\setlength{\tabcolsep}{5.1pt}
\caption{The ICC scores (with 95\% confidence intervals [95\%CI]) for Agatston score (AS) and the volume-weighted (AS weighted) Agatston score. Results computed on the test set ($\TestSet$) consisting of 122 CT scans with our GT.}\label{table:results_classes}
\scalebox{0.9}{\begin{tabular}{ccccccccccc}
\Xhline{2\arrayrulewidth}
 Metric   & &  Total AS&& LM& &LAD && LCx  && RCA \\
 \hline 
 AS &&1.00&&0.85&&0.99&&0.98&&1.00\\
(95\% CI)&&(1.00, 1.00)&&(0.8, 0.89)&&(0.98, 0.99)&&(0.98, 0.99)&&(1.00, 1.00)\\
 \hline
  AS weighted&&1.00&&0.84&&0.99&&0.98&&1.00\\
(95\% CI)&&(1.00, 1.00)&&(0.78, 0.88)&&(0.98, 0.99)&&(0.97, 0.99)&&(1.00, 1.00)\\

\Xhline{2\arrayrulewidth}
\end{tabular}}
\end{table}

In Fig.~\ref{table:results}, we gather the results obtained for the three test sets, and report the current state-of-the-art segmentation results, with Observer 1 being the best manual segmentation, and the 3$^{\rm rd}$ place indicating the best known automated algorithm developed within the orCaScore challenge prior to our submission. We can observe that our technique not only outperforms this method, and achieves inter-observer level accuracy, making it the best available, but also delivers competitive results on a broader test set, with manual GT segmentations generated in this study. To further justify the equivalence of the analysis conducted on our GT, we present the metric distributions using 24 NC CTs for which we had two types of GT in Fig.~\ref{table:results}(b--c). The proposed pipeline delivers precise calcification delineation, with the performance nearing (or exceeding) the inter-observer quality level. The outstanding performance of the proposed method on our test set can be further appreciated in Fig.~\ref{fig:as_corr_plot}, which shows that the total Agatston scores predicted by the algorithm are in almost perfect agreement with the GT (Pearson correlation of 0.9997).

\begin{figure}[ht!]
    \centering
    \newcommand{\mywidth}{0.33}
    \scalebox{0.93}{\begin{tabular}{ccc}
        (a) & (b) & (c)  \\
         \includegraphics[width=\mywidth\textwidth]{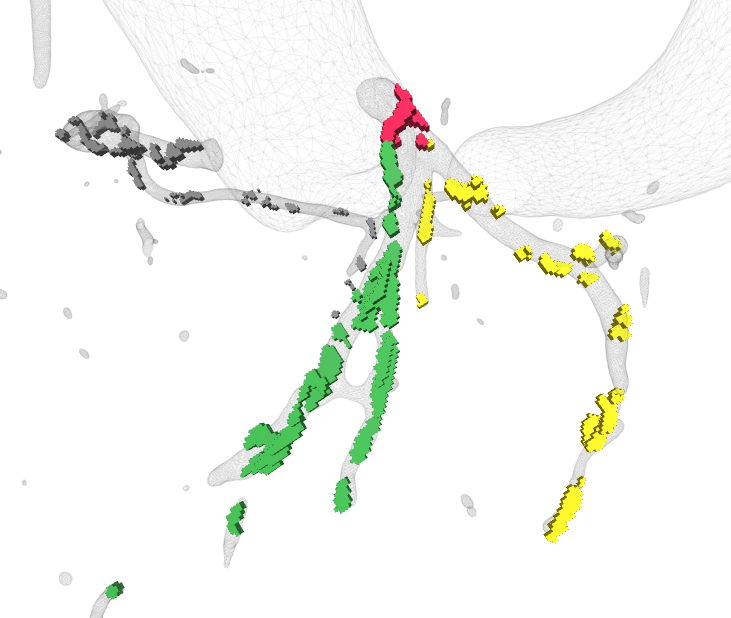}   & \includegraphics[width=\mywidth\textwidth]{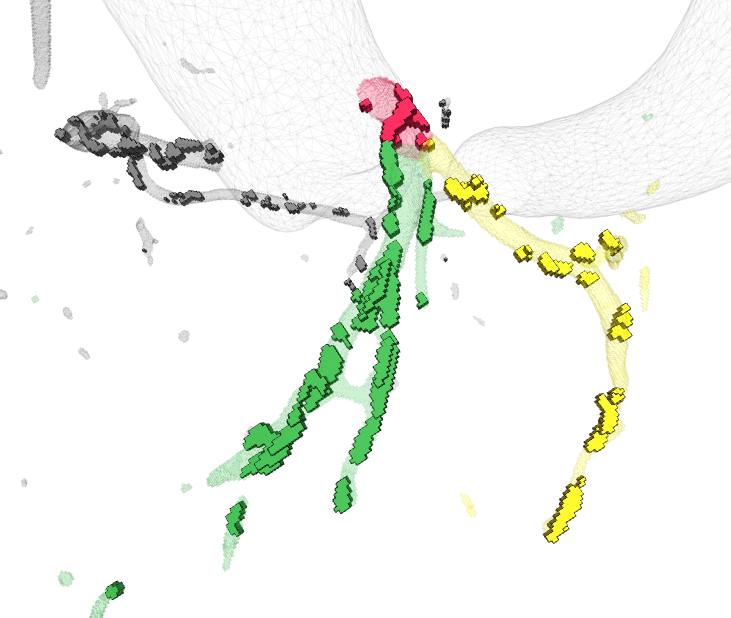} &  \includegraphics[width=\mywidth\textwidth]{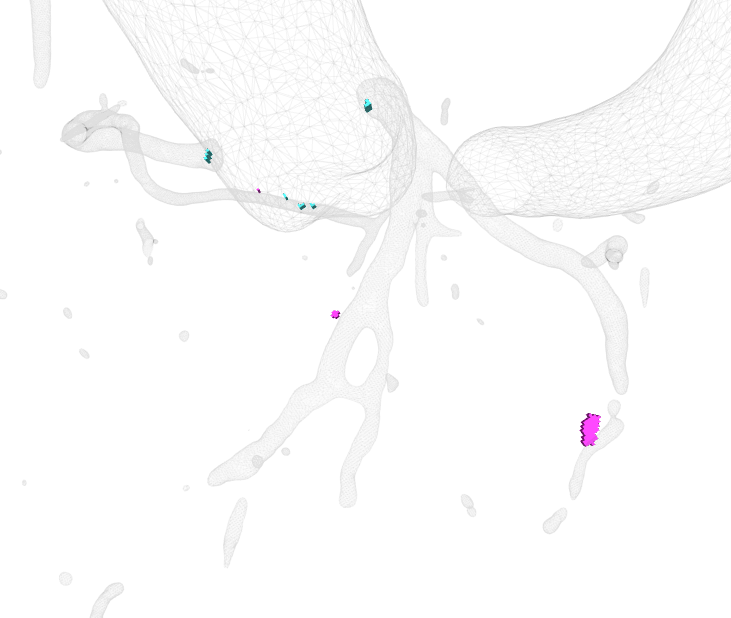}
    \end{tabular}}
    \caption{Visual results for the orCaScore test case with the highest Agatston score (1485~AU): (a) GT, (b) segmented calcifications assigned to four coronary vascular territories (indicated by different colors), and (c) false-positive (cyan) and false-negative (pink) detections. The DICE was as follows---total = 0.984, LM = 0.959, LAD = 0.973, LCx = 0.897, RCA = 0.979, and the relative AS difference = 1.91\%. }
    \label{fig:example}
\end{figure}

To investigate more deeply the proposed approach, for each element of $\TestSet$ (based on our GT), we calculated a four-class and five-class CVD risk category based on the Agatston score~\citep{hong2002coronary}---concordance for risk category assessment measured by the Cohen’s $\kappa$ for these patients was $0.97$ and $0.95$, respectively. Finally, Table~\ref{table:results_classes} presents a nearly perfect agreement between automatically and manually extracted Agatston scores, without and with weighting according to the calcifications' volumes, while Table~\ref{fig:classification_regions} shows key segmentation metrics for the four coronary artery regions considered in this work. The high quality of these results shows the potential clinical utility of the system, which allows to objectively calculate parameters (both at the entire scan level, and for specific parts of the vessel tree, without any human-related bias). Despite offering high-quality numerical operation, our approach enables practitioners further investigate the locations of calcifications, effectively enhancing the analysis process (Fig.~\ref{fig:example}). The segmentation result can be thus easily explained by attributing each calcification to a particular location along the coronary vessel graph---it is of paramount significance in clinical settings to design more precise treatment pathway.

\begin{table}[ht!]
\centering
\label{fig:classification_regions}
\caption{Segmentation performance metrics (mean, standard deviation, and calcification volume weighted mean) for four different coronary artery tree regions (LM, LAD, LCx, RCA) computed based on our test set ($\TestSet$) consisting of 122 CT scans with our GT.}
\vspace{3pt}
\scalebox{0.65}{\begin{tabular}{cccccccccccc}
\toprule
 & \multicolumn{3}{c}{DICE} & & \multicolumn{3}{c}{Recall} & & \multicolumn{3}{c}{Precision} \\
\cline{2-4} \cline{6-8} \cline{10-12}
CA tree region & Mean & Std. & Weighted & & Mean & Std. & Weighted & & Mean & Std. & Weighted \\
\hline
LM & 0.730 & 0.269 & 0.820 & & 0.814 & 0.255 & 0.868 & & 0.760 & 0.292 & 0.818 \\
LAD & 0.934 & 0.199 & 0.949 & & 0.940 & 0.197 & 0.944 & & 0.953 & 0.180 & 0.977 \\
LCx & 0.885 & 0.282 & 0.921 & & 0.899 & 0.266 & 0.902 & & 0.890 & 0.289 & 0.950 \\
RCA & 0.959 & 0.094 & 0.973 & & 0.979 & 0.051 & 0.969 & & 0.953 & 0.119 & 0.980 \\
\bottomrule
\end{tabular}}
\end{table}

\begin{table}[ht!]
\scriptsize
\centering
\setlength{\tabcolsep}{10pt}
\caption{Consecutive steps of the coronary artery calcium scoring process in four challenging scenarios. Cases with: (a) implanted pacemaker leads, (b) significant image noise, (c) aortic valve calcifications, and (d) mitral valve calcification. Voxels $\geq$~130~HU are shown in blue, and predicted coronary artery calcifications are marked in red.}\label{table:results_examples}
\vspace*{5pt}
\scalebox{0.9}{
\begin{tabular}{cccc}
\Xhline{1\arrayrulewidth}
 & Multi-organ segmentation &  HU $\geq$ 130 within pericardial sac & CA calcifications \\
 \hline 
\\
\vspace*{2pt} \raisebox{15ex}{(a)} & \includegraphics[width=0.25\textwidth]{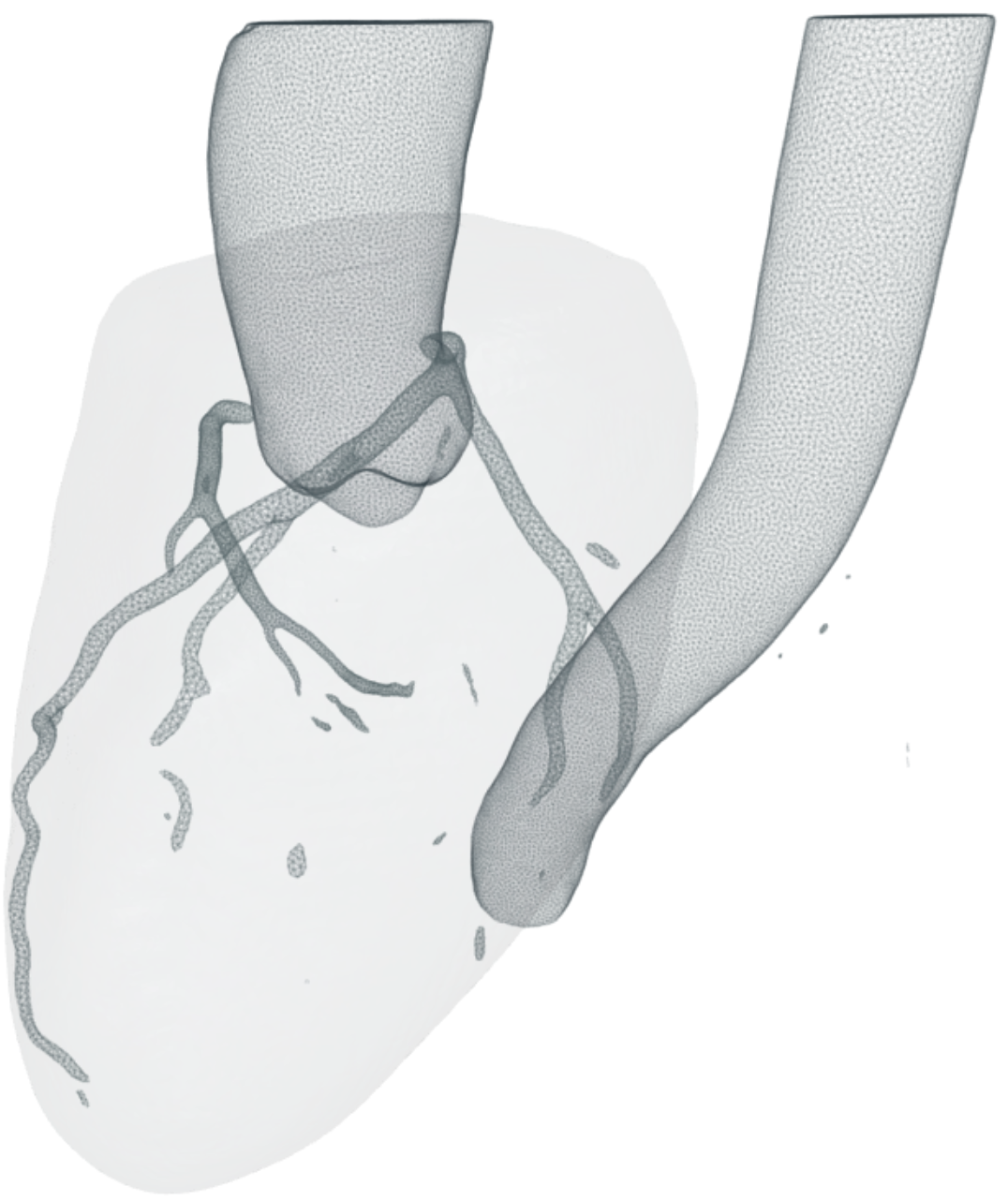} &  \vspace*{2pt}\includegraphics[width=0.25\textwidth]{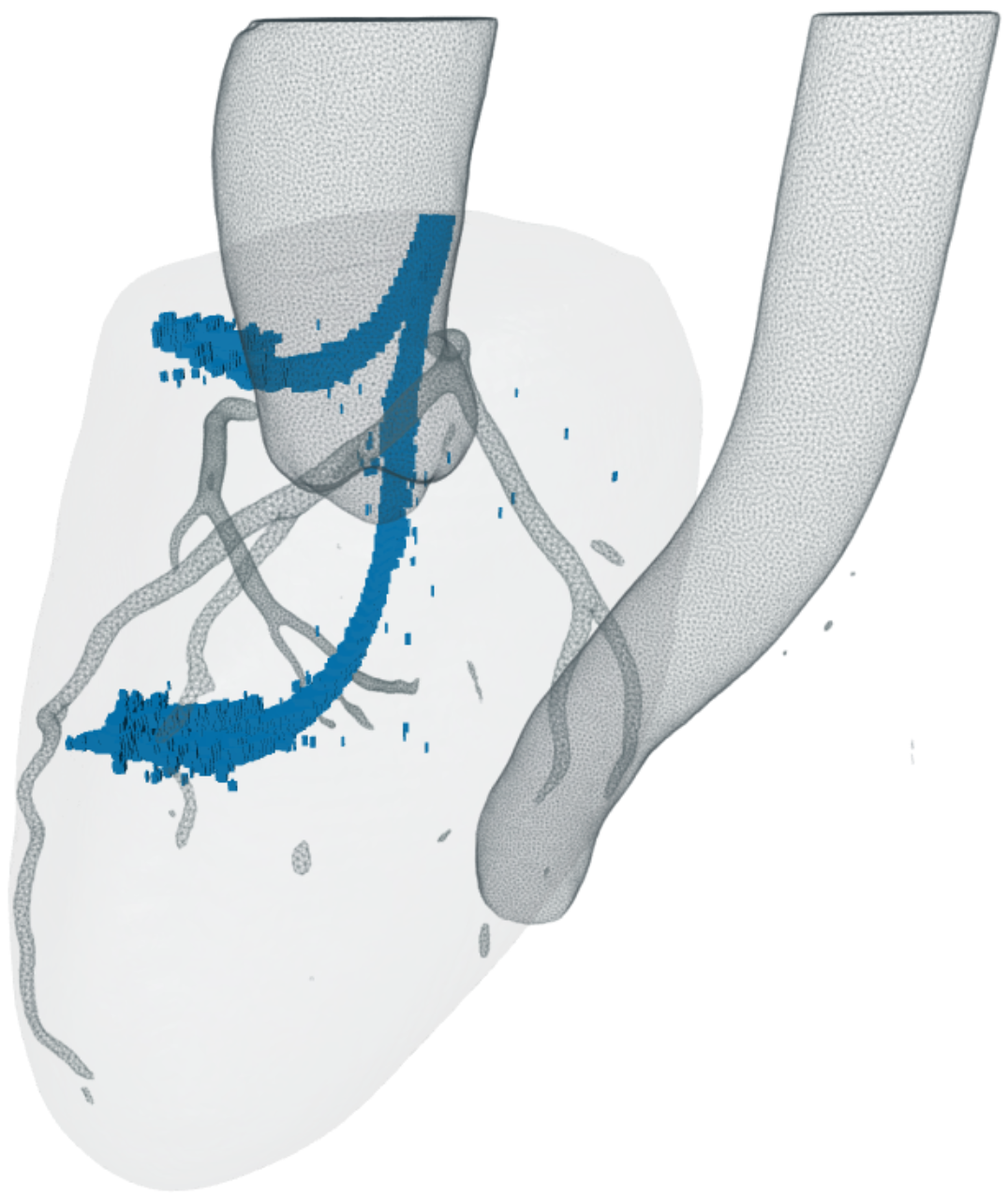} & \vspace*{2pt}\includegraphics[width=0.25\textwidth]{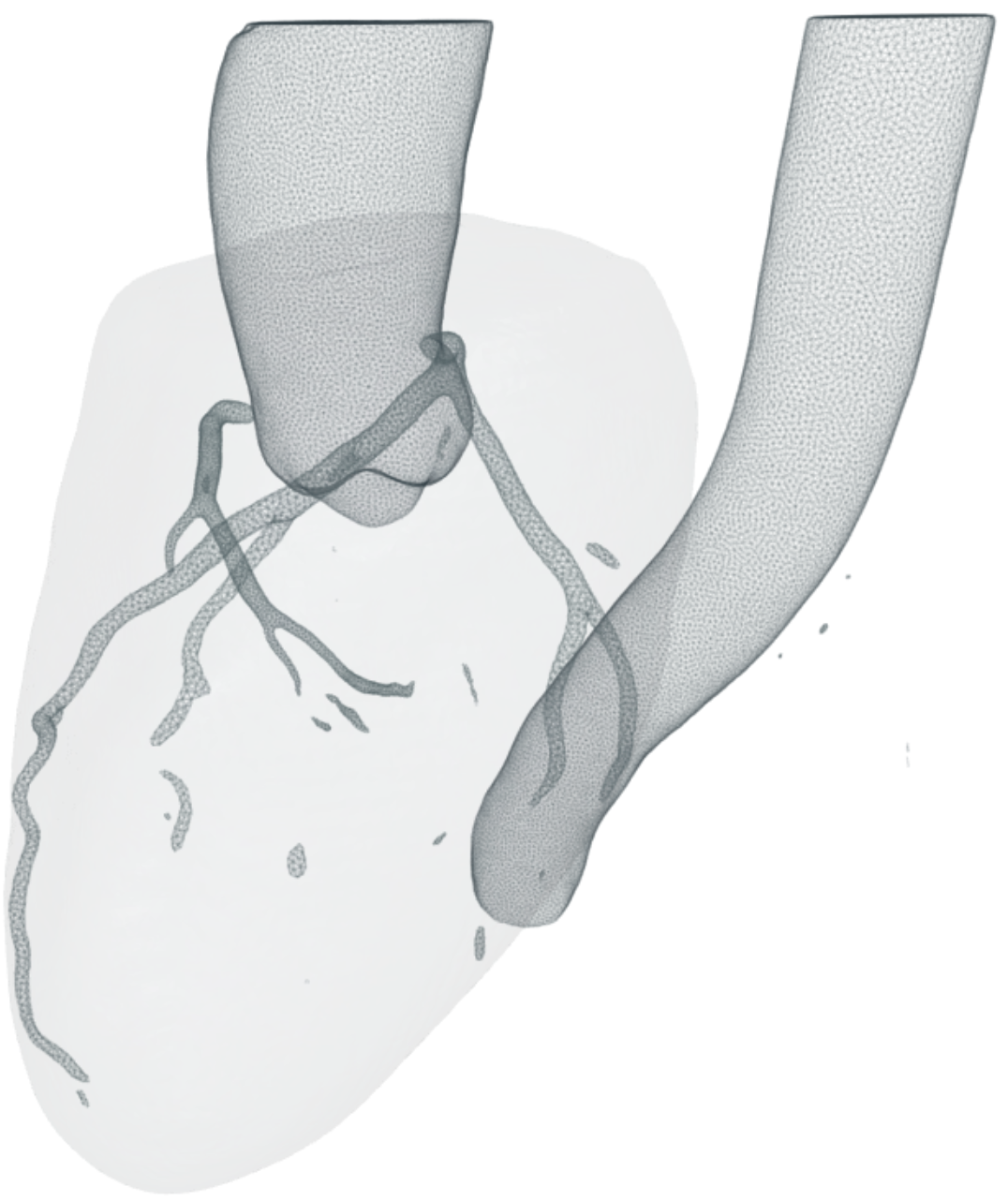} \\
 
 \hline 
\\
\vspace*{2pt} \raisebox{15ex}{(b)} & \includegraphics[width=0.25\textwidth]{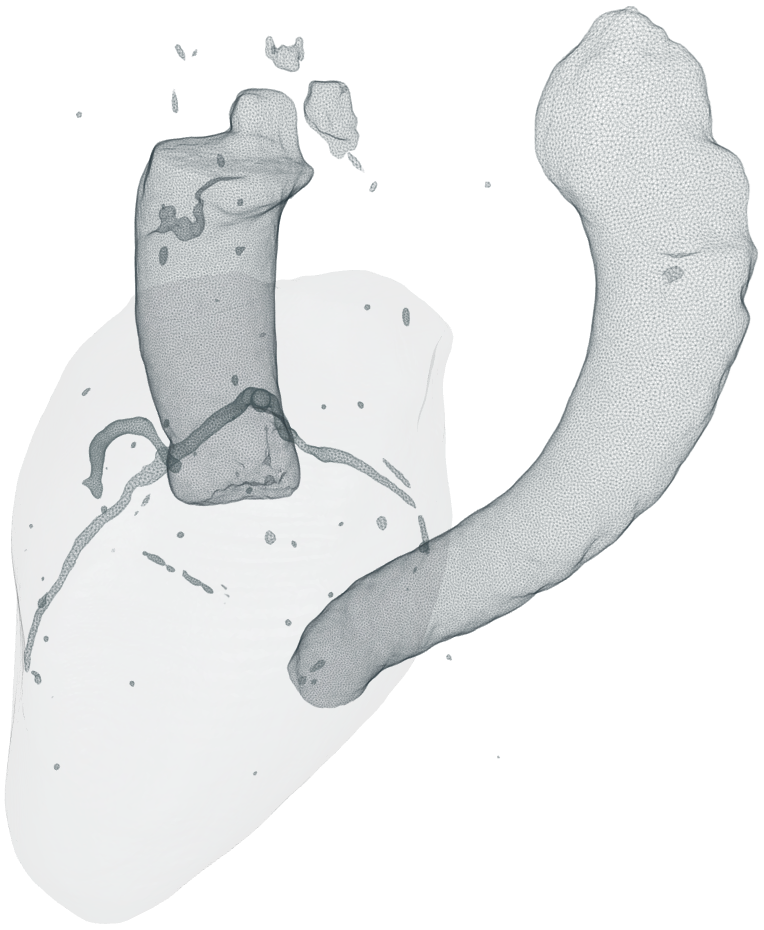} &  \vspace*{2pt}\includegraphics[width=0.25\textwidth]{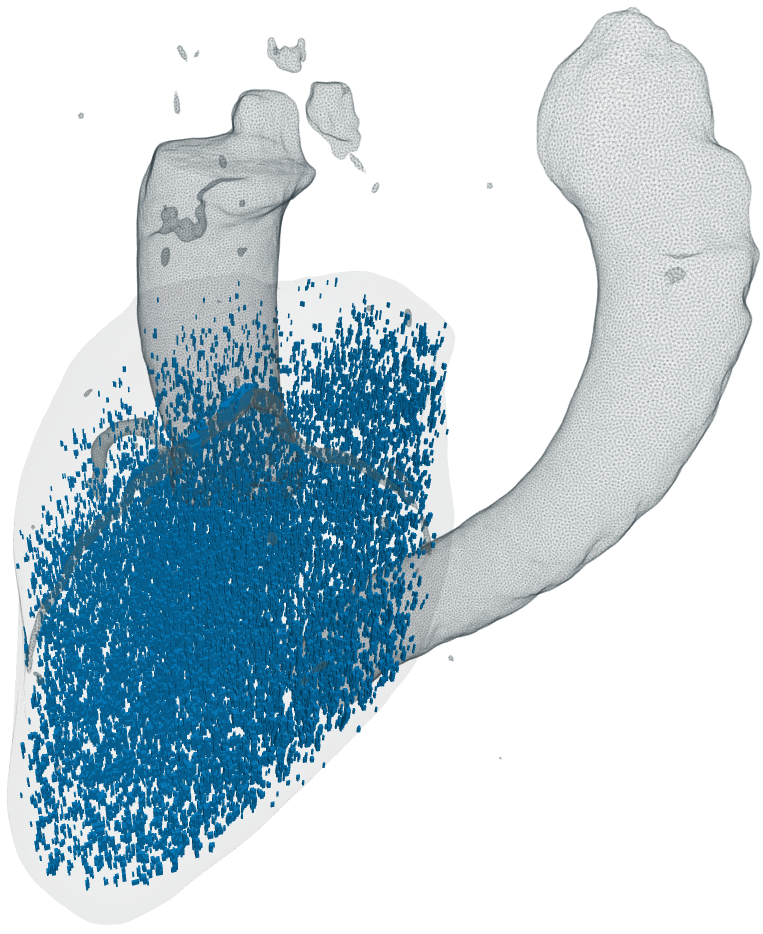} & \vspace*{2pt}\includegraphics[width=0.25\textwidth]{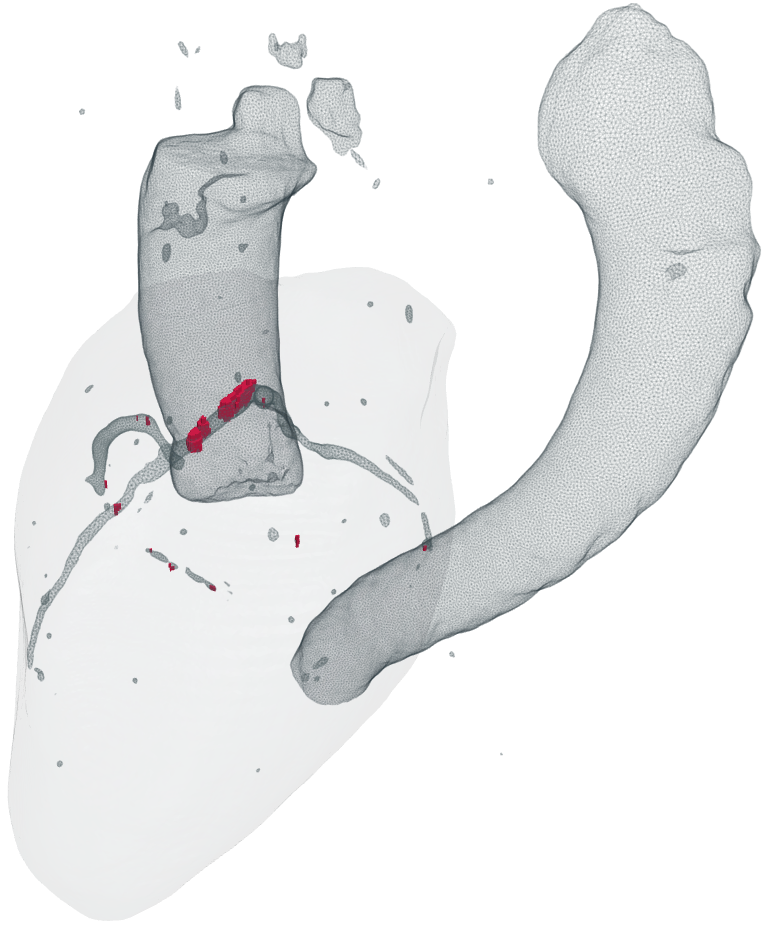} \\

 \hline 
\\
\vspace*{2pt} \raisebox{10ex}{(c)} & \includegraphics[width=0.25\textwidth]{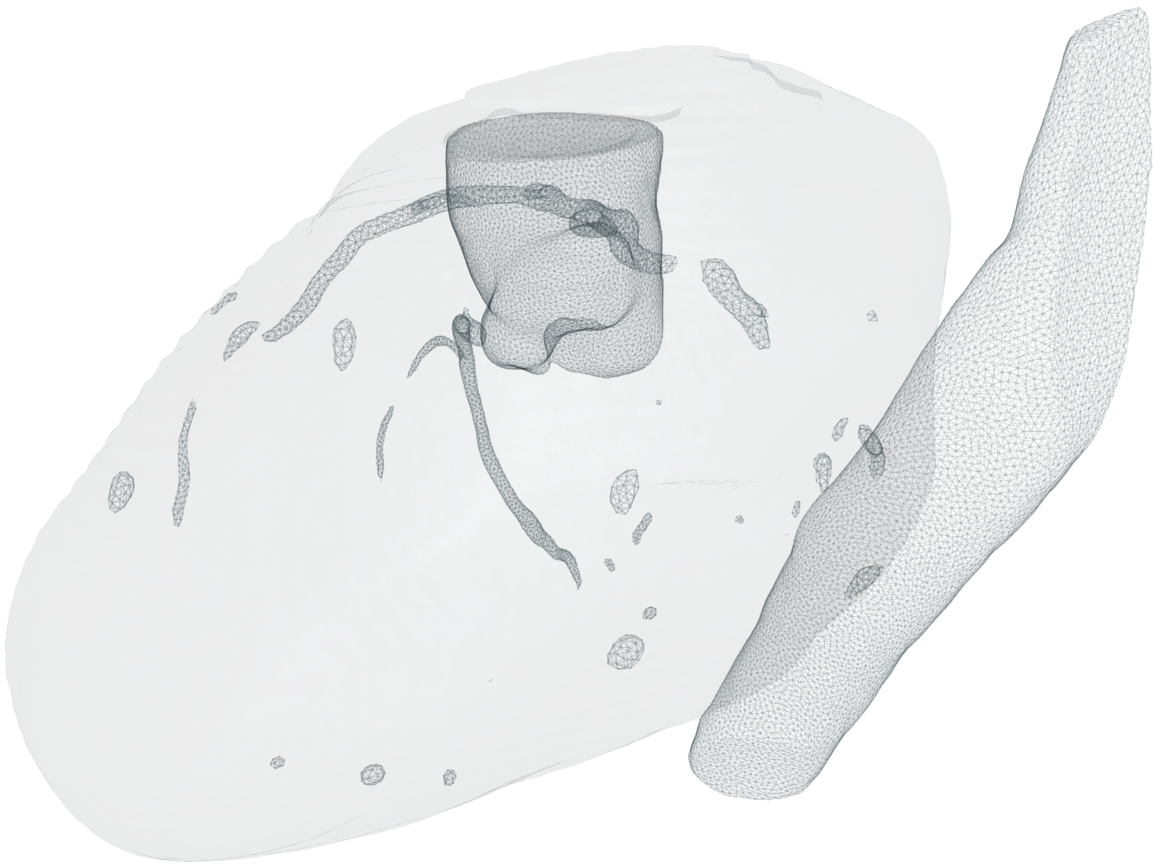} &  \vspace*{2pt}\includegraphics[width=0.25\textwidth]{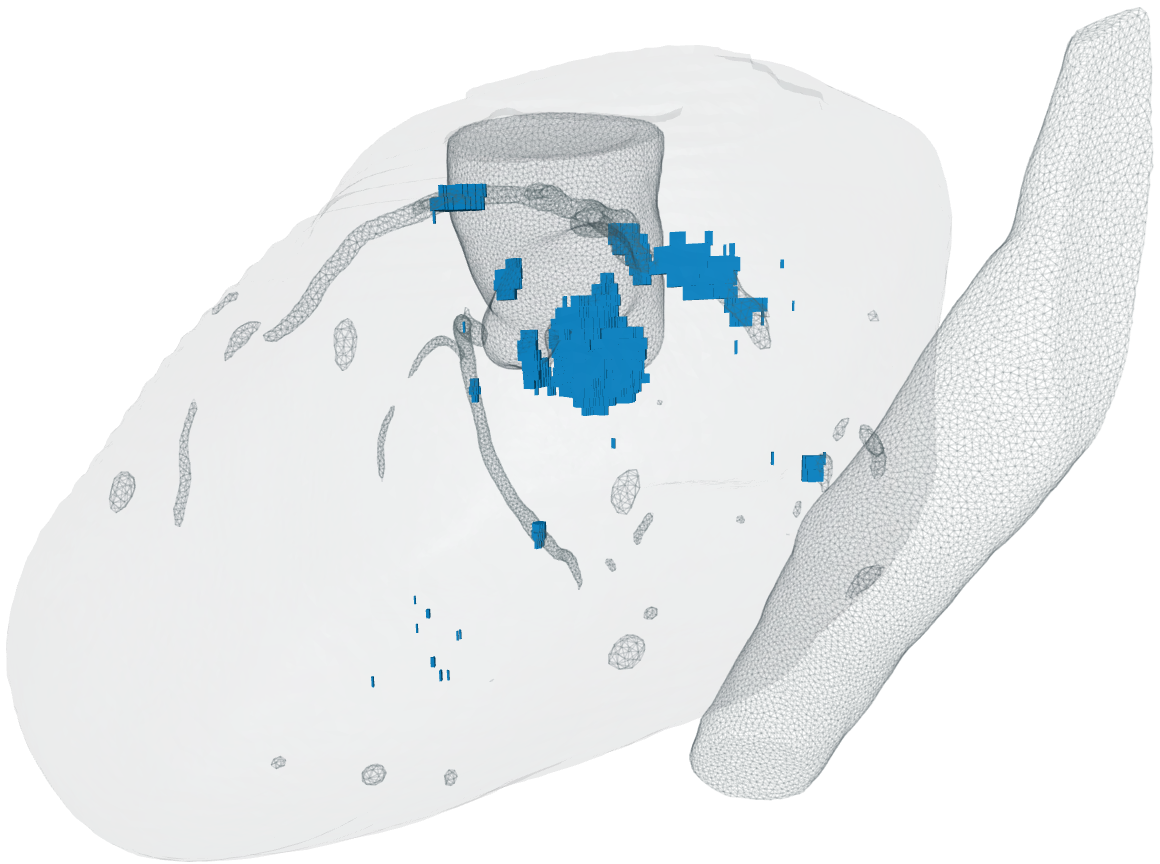} & \vspace*{2pt}\includegraphics[width=0.25\textwidth]{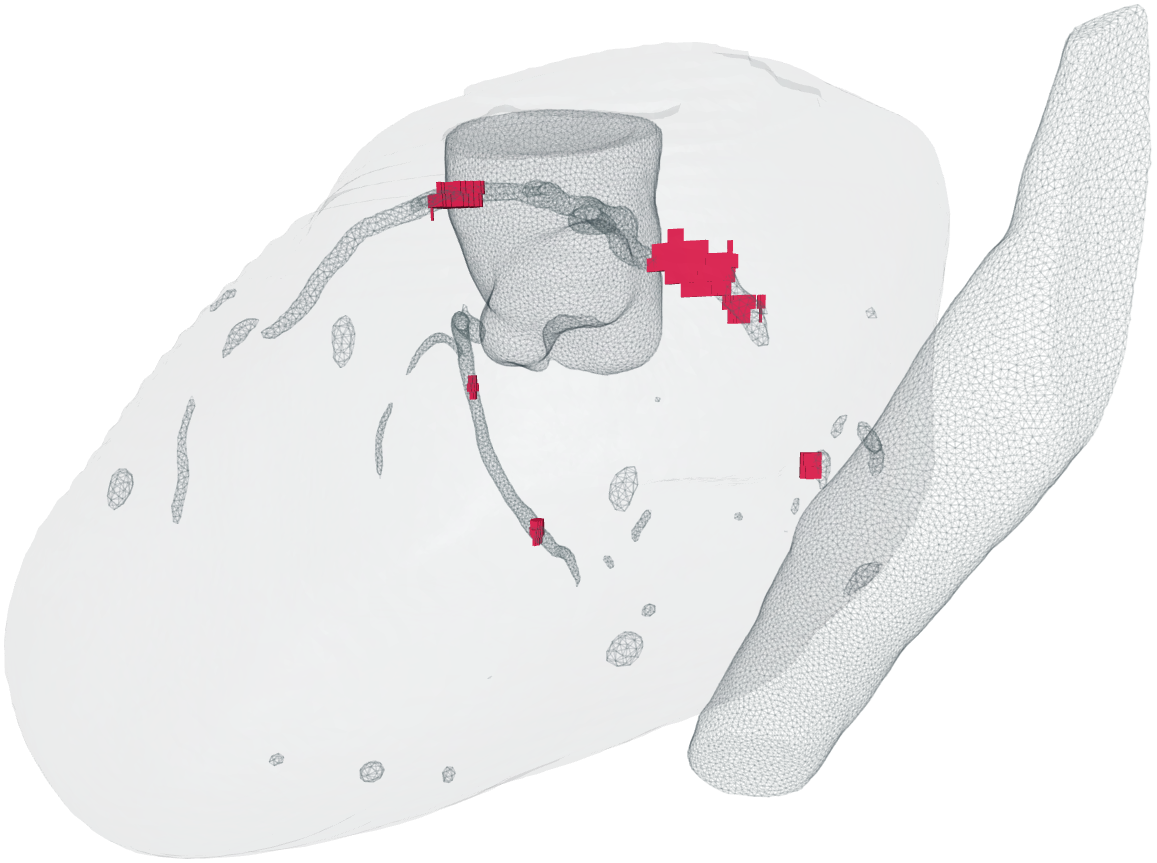} \\

 \hline 
\\
\vspace*{2pt} \raisebox{15ex}{(d)} & \includegraphics[width=0.25\textwidth]{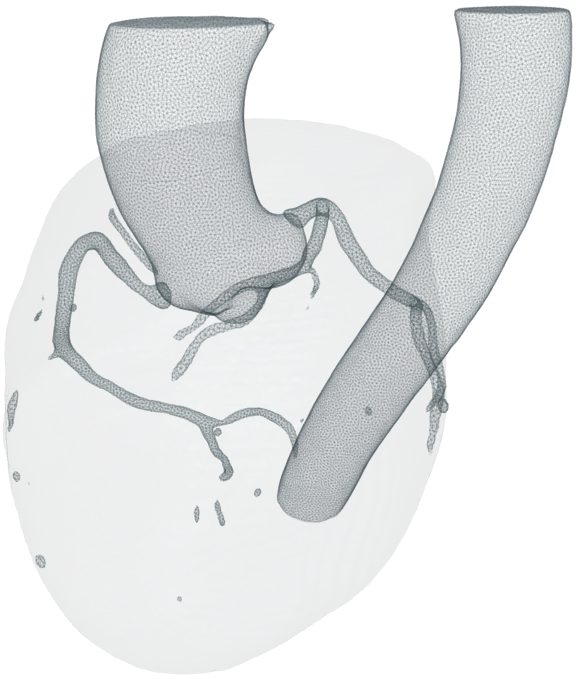} &  \vspace*{2pt}\includegraphics[width=0.25\textwidth]{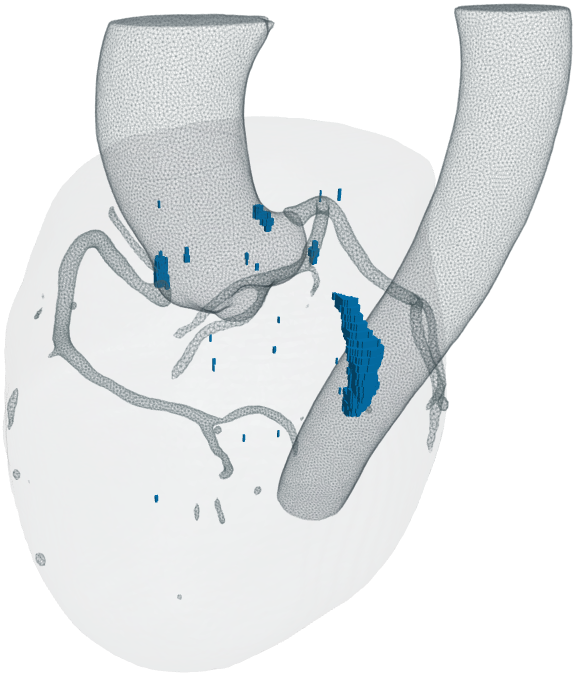} & \vspace*{2pt}\includegraphics[width=0.25\textwidth]{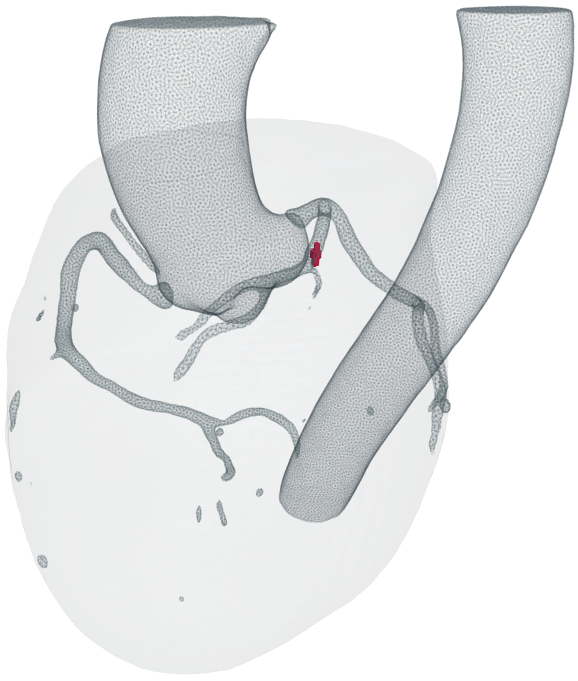} \\

\Xhline{1\arrayrulewidth}
\end{tabular}
}
\label{tab:alg_steps}
\end{table}

Studying the consecutive steps of the calcium scoring algorithm provides valuable insight. Table \ref{tab:alg_steps} presents intermediate processing steps of our method, i.e., organ segmentation, thresholding of voxels $\geq$ 130 HU, and identification of coronary artery calcifications, in 4 scenarios challenging for standard calcium scoring algorithm. By utilizing anatomical information, our approach successfully distinguishes coronary calcifications from metallic implants, CT scan noise, and aortic and mitral valve calcifications, further highlighting the importance of semantic understanding and explainability in the process.


\section{Conclusions}

Coronary artery calcium scoring using NC ECG-gated cardiac CT scans is key for assessing the cardiovascular disease risk. Manual analysis of of multi-vendor CTs is, however, time-consuming and prone to human bias. We tackled this issue and proposed an end-to-end approach for locating calcifications, assigning them to specific anatomical parts of the coronary artery tree, and quantifying calcium scores across different anatomical areas of the heart. Our method provides interpretable segmentation masks for convenient calcification investigation and score calculation, and offers fast operation. While our technique shows high segmentation performance and nearly perfect agreement with human experts, and it was verified over the significantly larger dataset of CTs than orCaScore, we still aim for multi-center validation and invite the community to assess our approach across a broad range of CT studies, covering multiple scanner models and representing diverse patient populations.

\section*{Funding}
This work was supported by the National Centre for Research and Development POIR.01.01.01-00-0092/20. Jakub Nalepa was supported by the Silesian University of Technology grant for maintaining and developing research potential.









\bibliographystyle{elsarticle-harv}
\bibliography{main}

\end{document}